\documentclass[11pt,a4paper]{article}
\usepackage[hyperref]{acl2021}
\usepackage{times}
\usepackage{latexsym}
\usepackage{booktabs}
\usepackage{color}

\usepackage{adjustbox}
\usepackage{multirow}
\usepackage{color, colortbl}
\definecolor{Gray}{gray}{0.93}

\definecolor{ForestGreen}{RGB}{34,139,34}
\newcommand{\Wcal}{\mathcal{W}}

\newcommand{\soloist}{\textsc{Soloist}}
\newcommand{\synergy}{\textsc{Synergy}}
\newcommand{\model}{\textsc{Synergy}}
\newcommand{\modelp}[1]{\soloist{$^{\mathtt{#1}}$}}

\newcommand{\longname}{\textsc{\textbf{SY}mbolic k\textbf{N}owledge \textbf{E}mpowe\textbf{R}ed dialo\textbf{G} s\textbf{Y}stems}. In this paper, \model{} refers to both the proposed framework and the dialog model or system developed using the method.}

\makeatletter
\newcommand{\thickhline}{%
    \noalign {\ifnum 0=`}\fi \hrule height 1pt
    \futurelet \reserved@a \@xhline
}
\makeatother

    \makeatletter
\def\@fnsymbol#1{\ensuremath{\ifcase#1\or \dagger\or \ddagger\or
   \mathsection\or \mathparagraph\or \|\or **\or \dagger\dagger
   \or \ddagger\ddagger \else\@ctrerr\fi}}
    \makeatother

\usepackage{microtype}

\usepackage{amsmath}
\usepackage{amsfonts}
\usepackage{amssymb}
\usepackage{wrapfig}
\usepackage{subcaption}
\usepackage{multirow}
 \usepackage{mathtools} 

\usepackage{verbatim}

\usepackage{anyfontsize}

\usepackage{color}
\usepackage{tikz}
\usetikzlibrary{arrows,shapes,snakes,automata,backgrounds,fit,petri}
\usepackage{adjustbox}

\newcommand{\RN}[1]{%
	\textup{\lowercase\expandafter{\it \romannumeral#1}}%
}

\usepackage{booktabs}

\usepackage{tcolorbox}

\makeatletter
\newcommand{\distas}[1]{\mathbin{\overset{#1}{\kern\z@\sim}}}%

\usepackage{enumitem}

\usepackage[lined,boxed,commentsnumbered,ruled,linesnumbered]{algorithm2e}

\newcommand{\ie}[0]{\emph{i.e., }}

\newcommand{\eg}[0]{\emph{e.g., }}

\newcommand{\beq}{\vspace{0mm}\begin{equation}}
\newcommand{\eeq}{\vspace{0mm}\end{equation}}
\newcommand{\beqs}{\vspace{0mm}\begin{eqnarray}}
\newcommand{\eeqs}{\vspace{0mm}\end{eqnarray}}
\newcommand{\barr}{\begin{array}}
\newcommand{\earr}{\end{array}}

\newcommand{\Ccal}{\mathcal{C}}
\newcommand{\Dcal}{\mathcal{D}}
\newcommand{\Fcal}{\mathcal{F}}
\newcommand{\Tcal}{\mathcal{T}}

\newcommand{\Scal}{\mathcal{S}}
\newcommand{\Hcal}{\mathcal{H}}

\ifx\assumption\undefined
\newtheorem{assumption}{Assumption}
\fi

\ifx\definition\undefined
\newtheorem{definition}{Definition}
\fi

\ifx\remark\undefined
\newtheorem{remark}{Remark}
\fi

\aclfinalcopy 
 
\begin{document}
%
\title{\synergy: Building Task Bots at Scale Using Symbolic Knowledge and Machine Teaching}

\author{Baolin Peng$^1$, Chunyuan Li$^1$, Zhu Zhang$^{12}$\thanks{~~Work was done when Zhu Zhang was visiting MSR}~, Jinchao Li$^1$, Chenguang Zhu$^1$, Jianfeng Gao$^1$ \\
  $^1$Microsoft Research, Redmond, WA \\
  $^2$Iowa State University / Ames, IA \\
  \texttt{\{bapeng,chunyl,jincli,chezhu,jfgao\}@microsoft.com} \\
  \texttt{zhuzhang@iastate.edu}
 }

\maketitle
\begin{abstract}
In this paper we explore the use of symbolic knowledge and machine teaching to reduce  human data labeling efforts in building neural task bots. 
We propose \model{\footnote{\longname{}}}, a hybrid learning framework where a task bot is developed in two steps:
$(\RN{1})$ Symbolic knowledge to neural networks: Large amounts of \emph{simulated} dialog sessions are generated based on task-specific symbolic knowledge which is represented as a task schema consisting of dialog flows and task-oriented databases. Then a pre-trained neural dialog model, \soloist{}~\cite{peng2020soloist}, is fine-tuned on the simulated dialogs to build a bot for the task.
$(\RN{2})$ Neural learning: The fine-tuned neural dialog model is continually refined with a handful of \emph{real} task-specific dialogs via machine teaching, where training samples are generated by human teachers interacting with the task bot.
We validate \model{} on four dialog tasks. 
Experimental results show that 
\model{} maps task-specific knowledge into neural dialog models achieving greater diversity and coverage of dialog flows, and continually improves model performance with machine teaching, thus demonstrating
strong synergistic effects of symbolic knowledge and machine teaching.
\end{abstract}

\section{Introduction}

\begin{figure*}[t!]
\centering
\includegraphics[width=2\columnwidth]{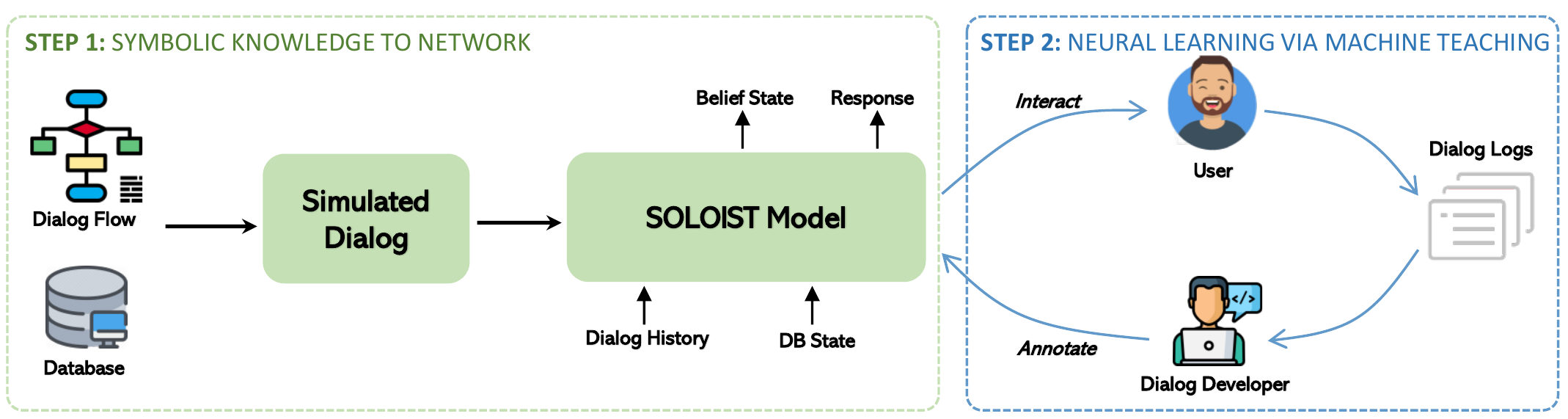}
\caption{The proposed \model{} framework: {\color{ForestGreen}Step 1}, the dialog developer provides task schema that includes dialog flows and task-specific database, simulators generate simulated dialogs, and an end-to-end model is trained using the simulated data. {\color{blue}Step 2},  the model is refined using machine teaching.}
\label{fig:flow}
\end{figure*}

Recent advances in the design and training  of deep neural networks have led to great success in building task-oriented dialog systems \cite{wen2016network,peng2017composite,zhou2018design,gao2019dialog,gao2020robust,peng2020soloist,Wu2020ToDBERTPN}. Though demonstrably effective, building such neural dialog systems typically requires an extensive collection of dialogs in the target task domain to achieve good generalization performance, even when the task can be described concisely in symbolic representations \cite{shah2018building,goel2019hyst,eric2020multiwoz}. 

Crowd-sourcing is one of the most popular approaches to collecting and annotating dialog corpora \cite{budzianowski2018multiwoz,eric2020multiwoz}. 
To generate a dialog session, two workers play the roles of user and agent, respectively. 
The user initiates a dialog according to a pre-set user goal. 
Then, the agent converses with the user to generate the dialog session by accomplishing the goal. 
However, this process suffers from several potential issues 
$(\RN{1})$ it is time-consuming and expensive to collect a large enough corpus to cover various dialog samples.
$(\RN{2})$ the annotations of intents and dialog states inevitably contain errors, incurring non-trivial post-processing cost to ensure data quality. 
More importantly, most exiting task bot building methods are based on training neural dialog models on conversational data \cite{gao2019neural}, but do not explicitly leverage task-specific symbolic knowledge that is often more compact and expressive than raw data.
Therefore, there has been a surge of interest in combining symbolic knowledge (e.g., task schema) and deep learning to build dialog systems \citep[e.g.][]{williams2017hybrid,shukla2020conversation,gao2020robust}. 
For task-oriented dialog scenarios, {\it task schema} defines every aspect of a task at the meta level. It contains $(\RN{1})$ {\it dialog flows} that are represented by graphs and describe anticipated interaction patterns 
\footnote{Many popular commercial tools for task bot building allow dialog authors to compose dialog flows using domain knowledge, including Google's Dialog Flow, Microsoft's Power Virtual Agent, Facebook's Wit.ai and Amazon's Lex. Although complete dialog flows are hard to compose except for very simple dialog tasks, they represent some of the most representative human-machine interaction patterns. As we will show in this paper, dialog flows can be leveraged to build neural dialog models.}
;
$(\RN{2})$ {\it task-specific databases} that contain all the slot and values that a dialog system need to handle. 



In this paper, we propose \model{}, a hybrid learning framework to exploit symbolic knowledge and machine teaching to rapidly build task bots with minimal human annotation efforts. 
As shown in Figure \ref{fig:flow}, \model{} builds a task bot in two steps. 
(1) Translating symbolic knowledge to neural networks. A dialog developer provides task schema that contains dialog flows and task-specific databases. Then, large amounts of simulated dialogs are generated by traversing the dialog flows and enumerating data records in the databases. Lastly, the task bot is built by fine-tuning the pre-trained neural dialog model \soloist{}~\cite{peng2020soloist} on the simulated dialogs. 
Despite its simplicity, Step 1 cannot guarantee completeness for real-world tasks in that the dialog model is only trained on simulated dialogs derived from task schema. The model needs to be refined using real dialogs in Step 2.
(2) Neural learning: The fine-tuned neural dialog model is continually refined with a handful of real dialogs via machine teaching, where training dialogs are generated by human teachers interacting with the task bot. 
%
\model{} renders synergistic combinations of symbolic knowledge and deep learning, including 
$(\RN{1})$ effortlessly cover all interaction patterns in the dialog flows; 
$(\RN{2})$ cover all possible slot values to avoid unseen value extraction issues, which often occurs for commercial systems; 
$(\RN{3})$ correctness of supervised labels for model training; 
$(\RN{4})$ introduction of machine teaching to continually refine the model. 
Our experiments on four domains show that incorporating symbolic knowledge brings substantial improvement on task-completion and dialog state tracking and results in a good starting point for machine teaching. 

To sum up, we make the following contributions:
\begin{itemize}
  \item We are the first to exploit symbolic knowledge (\ie task schema) to reduce the cost of building task bots at scale.
  \item We propose \model{}, a hybrid learning framework that can rapidly build task bots with minimal human annotation efforts by using symbolic knowledge and machine teaching.
  \item We demonstrate that \model{} can achieve state-of-the-art performance with only 10\% human annotation cost, compared to previous SOTA methods, on four well-studied dialog tasks.
\end{itemize}

\section{Methodology}
\subsection{Motivation}

Conceptually, there are two different approaches to building intelligent systems~\cite{towell1994knowledge,kambhampati2021polanyi}: {\it symbolic knowledge} and {\it empirical learning}. Suppose our goal is object classification.
The symbolic knowledge approach leverages ``domain knowledge'' to identify and recognize critical
facets of class members, and define rule-based hand-built classifier to distinguish objects. The empirical learning approach instead shows lots of training examples to a learner without any explanation of why the examples are members
of a particular class. After seeing sufficient examples, the learner is expected to ``discover'' the underlying concepts and classify new examples. Deep Learning is an effective method for empirical learning.

Unfortunately, neither of these two approaches to machine intelligence is completely satisfactory~\cite{towell1994knowledge}. For example, hand-built classifiers assume that their domain knowledge is complete and correct, which is often extremely difficult to achieve in most real-world tasks; On the other hand, deep learning models require a large number of examples to generalize well, and their initial parameters can greatly affect how well concepts are learned. Hence, hybrid learning methods are developed to use symbolic knowledge of a domain and a set of classified examples~\cite{towell1994knowledge}. These methods show encouraging performance on simple tasks such as promoter recognition (a special DNA sequence). It remains unknown how to adopt a similar idea in more complex scenarios such as task-oriented dialog systems.

Let us first review a traditional task-oriented dialog system, which has four modules and executes sequentially~\cite{DBLP:journals/pieee/YoungGTW13,gao2019neural}.
A natural language understanding ({NLU}) module identifies user intents and extracts associated information such as slots and their values from user’s input. A dialog state tracker ({DST}) infers the belief state (or user goal) from dialog history.
The belief state is often used to query a task-specific database (DB) to obtain the DB state, such as the number of entities that match the user goal.
The dialog state and DB state are then passed to a dialog policy ({POL}) to select the next system action.
A natural language generation ({NLG}) module converts the action to a natural language response. 
Recent research shows that the aforementioned modular pipeline can be 
formulated as a single auto-regressive model, such as \soloist{}~\cite{peng2020soloist} and SimpleTOD~\cite{hosseini2020simple}. 
In this study, we develop \model{} based on a pre-trained \soloist{} system.

The training data for fine-tuning the pre-trained \soloist{} model is a set of dialog sessions annotated with grounding information, \ie user goals, dialog belief states, database states, and system responses. All these items can be concatenated into a sequence of tokens, and fed into neural networks for model training. The bottleneck for scaling up training is how to easily collect such long sequences for different domains. In a traditional dialog development process, human efforts are required to engineer every aspect of the conversational interaction and anticipate all possible interaction patterns for completing the task, for example, annotating latent dialog states from observed natural language utterances. 
This renders the deployment of dialog systems less affordable in a wider range of domains.
Therefore, it is highly desirable to expand this approach by automating the dialog session collection process and bringing it closer to minimal-human-effort regimes. Our solution adopts the hybrid learning approach: we use domain knowledge to synthesize dialog sessions as training data and further refine the model via machine teaching.

%


\subsection{\synergy{}}

The proposed {\synergy{}} framework has two main components: symbolic knowledge to neural networks, and neural learning via machine teaching.

\subsubsection{Symbolic Knowledge to Neural Networks}
At a high level, \model{} starts with a task schema that contains dialog flows and task-specific databases in a target domain. Formally, the framework maps task schema $\Tcal$ to a set of $N$ dialog sessions $\Dcal$ in natural language:
\begin{align}
\vspace{-2mm}
\label{eq_high_level}
\Fcal ( \Tcal ) \rightarrow \Dcal = \{ d_i, i=1...N\} \\
d_i = (u_1^i, a_1^i, y_1^i, \cdots, u_n^i, a_n^i, y_n^i)
\vspace{-2mm}
\end{align}
where $d_i$ is a dialog session that comprises a sequence of utterances and its associated labels. For example, $u_1^i$ and $a_1^i$ are utterances from user and agent, respectively, $y_1^i$ is the dialog state accumulated till that turn. 

\begin{figure*}[hbt!]
\centering
\includegraphics[width=1.99\columnwidth]{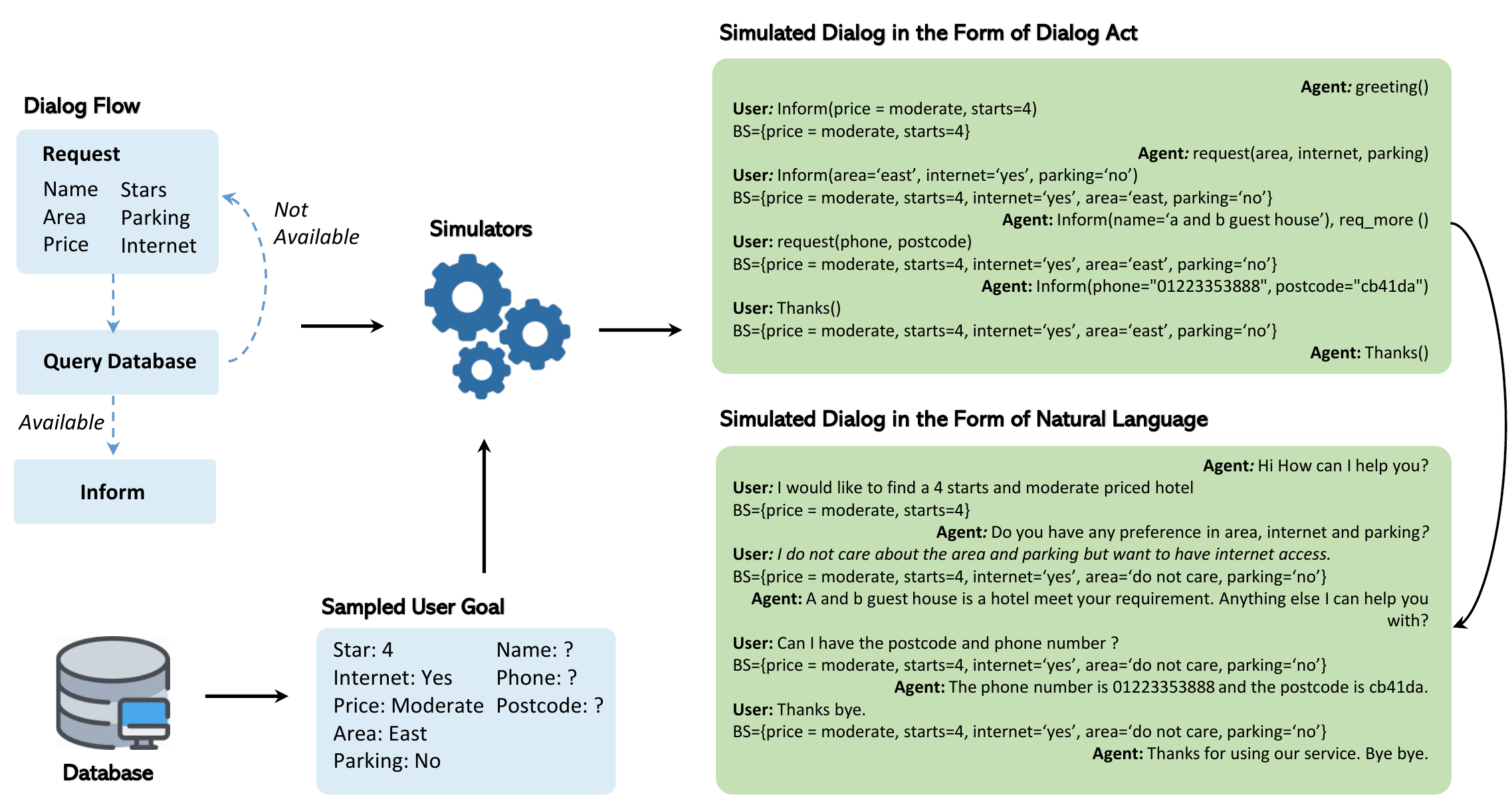}
\caption{An example of generating simulated dialogs in the Hotel domain. Dialog Flow defines the anticipated interaction patterns. Each function block in the dialog flow is implemented with agenda-based simulators. The database contains all possible values. For each dialog, a user goal is sampled according to a function block in the dialog flow and database. In what follows, agenda-based simulators generate dialogs in dialog act level, and rule-based NLG converts dialog acts into natural language.}
\label{fig:example}
\vspace{-1mm}
\end{figure*}

\paragraph{Task Schema.} 
Given a domain, task schema $\Tcal$ defines the scope of the dialog interactions for the task. In this paper, we focus on database querying applications, which involve entities that a user would like to browse and select through a natural language dialog. Note that the task schema is the only artifact (instead of labeled data) we request from the dialog developers.

Specifically, the task schema is formed by $\Tcal = (\Wcal, \Ccal)$, where $\Wcal$ is dialog flow and $\Ccal$ a database API. Dialog flow $\Wcal$ describes the interaction patterns and business rules to achieve the task.  The columns of the database $\Ccal$ are denoted as ``slots''. Each slot could be a constraint that the user cares about when selecting an entity or the information that the user intends to obtain. Further, the developer must provide a database API that can be queried with a SQL-like syntax to return a list of matching candidate entities for any valid combination of slot-value pairs. Figure \ref{fig:example} illustrates an example in the hotel domain. The task is to serve as an information desk to provide hotel information for users. The dialog flow describes the steps required to achieve the task, \ie 1) request necessary slots from users, 2) query the database based on acquired information from users. A business rule is introduced to inform users if entities match or ask users to start over.

\paragraph{Simulated Dialog Generation.}

Given the task schema, we generate a set of simulated dialogs centered around the task in two steps as: 
\begin{align}
\vspace{-2mm}
\Fcal ( \Tcal ) = \Hcal_2 ( \Hcal_1 ( \Tcal ) ) \\
\Hcal_1 ( \Tcal ) \rightarrow \Scal = \{ s_i, i=1...M \} \\
s_i = (\hat{u}_1^i, \hat{a}_1^i, \hat{y}_1^i, \cdots, \hat{u}_n^i, \hat{a}_n^i, \hat{y}_n^i) \\
\Hcal_2 ( \Scal ) \rightarrow \Dcal = \{ d_i, i=1...N \} \\
d_i = (u_1^i, a_1^i, y_1^i, \cdots, u_n^i, a_n^i, y_n^i)
\vspace{-2mm}
\end{align}
where the first step $\Hcal_1$ maps the task schema to a set of simulated dialog sketches $\Scal$, each being represented by a sequence of dialog acts $\hat{a}$, $\hat{u}$ and belief states $\hat{y}$, and the second step $\Hcal_2$ maps each dialog sketch $s$ to a dialog in natural language $d$.

The simulated dialog sketches summarize the dialog flow. 
Compared to dialogs in natural language, they can be easier generate using rule-based user simulators and agents, since they are in the form of dialog acts instead of complex and diverse natural language utterances. To generate a sketch, as shown in Figure \ref{fig:example}, \model{} firstly samples a user goal $g$ based on the constraints in the dialog flow and the database. A user goal in task-oriented dialog describes the objective that the user wants to achieve. Each user goal comprises informable slots that serve as constraints for database queries and requestable slots that the user wants to acquire from the dialog system. For example, the first function block of the dialog flow in Figure \ref{fig:example} is requesting from the user $\mathtt{Name}$, $\mathtt{Area}$, $\mathtt{Price}$, $\mathtt{Stars}$, $\mathtt{Parking}$, $\mathtt{Internet}$. Pursuant to this function and the database, a user goal is sampled where each informable slot is assigned a fixed value, and each requestable slot has a question mark to indicate that it is requestable. The values in a user goal are chosen by exhaustively enumerating the items in the database.

Given the sampled user goal and dialog flow, \model{} conducts self-play with a user simulator $\mathit{U}$ and an dialog agent simulator $\mathit{A}$ to generate simulated dialog sketches $s_i$:
\begin{align}
\vspace{-2mm}
\mathit{U} = P(\hat{u}_{j+1}^i, \hat{y}_{j+1}^i|\hat{u}_1^i, \hat{a}_1^i, \hat{y}_1^i, \cdots, \hat{u}_j^i, \hat{a}_j^i, \hat{y}_j^i, g_i) \\ 
\mathit{A} = P(\hat{a}_{j+1}^i|\hat{u}_1^i, \hat{a}_1^i, \hat{y}_1^i, \cdots, \hat{u}_{j+1}^i, g_i, \Tcal)
\vspace{-2mm}
\end{align}
where $\hat{u}_{j}^i$ denotes the dialog act, represented by intent and slot-values, of user simulator $\mathit{U}$ at the $j$-th turn, and $\hat{y}_{j+1}^i$ means the dialog state. For example, in Figure \ref{fig:example}, $\mathtt{inform(price=moderate, starts=4)}$ is to inform the user's preference on price and starts for the searching. $\mathit{U}$ models the distribution of next possible user dialog acts and dialog states given dialog history and user goals. Similarly, $\mathit{A}$ maps dialog history and task schema $\Tcal$ to the next most appropriate dialog system action. 
There are several options to implement $\mathit{U}$ and $\mathit{A}$. For simplification, in this paper, we use task-independent agenda-based simulators \cite{schatzmann2007agenda} for both $\mathit{U}$ and $\mathit{A}$. By exploring a large number of user goals using dialog flows and databases, numerous and diverse simulated dialog sketches can be obtained. 

The second step $\Hcal_2$ maps the sketches $\Scal$ to dialogs in natural language $\Dcal$. This can be achieved using semantically conditioned natural language generation model \cite{peng2020few}. 
For example, $\mathtt{inform(price=moderate, starts=4)}$ can be converted to ``The price is moderate. the start is 4". Alternatively, dialog developers can provide more informative templates to improve user experience.

\paragraph{Model Training with Simulated Dialogs.}
In our framework, any end-to-end neural dialog models can be trained on the simulated dialogs $\Dcal$. 
In this paper we fine-tune the pre-trained \soloist{} model, denoted as $\mathcal{M}$, on $\Dcal$ since \soloist{} has shown record performance on a wide range of dialog tasks and requires low annotation cost. 

\subsubsection{Neural Learning via Machine Teaching}

$\mathcal{M}$ trained on the simulated dialogs $\Dcal$ gains the capability of chatting like humans to complete the task. It can be further refined using a handful of real dialogs. To achieve this goal, we employ Conversation Learner~\cite{shukla2020conversation}, an effective machine learning tool that allows human teachers (dialog authors) to visualize logged human-system dialogs, find potential problems, provide corrections or additional examples to improve systems' performance. 
It operates in the following steps: 
$(\RN{1})$ Dialog authors deploy $\mathcal{M}$ for a specific task. 
$(\RN{2})$ Dialog developers or users interact with the deployed system and generate human-system dialog logs.
$(\RN{3})$ Dialog authors revise a handful of training samples by selecting representative failed dialogs, correcting their belief states and responses so that the system can complete these dialogs successfully. 
The corrected dialogs are then used to fine-tune model $\mathcal{M}$.

\begin{table*}[htbp]
    \centering
    \small
    \scalebox{0.8}{
    \setlength{\tabcolsep}{1.0mm}{
    \begin{tabular}{lcccccccccccccccc}
    \toprule
    
    \multirow{2}{*}{Model} & \multirow{2}{*}{\#Example} &
\multicolumn{3}{c}{\texttt{Attraction}} &
\multicolumn{3}{c}{\texttt{Train}} &
\multicolumn{3}{c}{\texttt{Hotel}} &
\multicolumn{3}{c}{\texttt{Restaurant}} \\
\cmidrule(l){3-5} \cmidrule(l){6-8} \cmidrule(l){9-11} \cmidrule(l){12-14} \cmidrule(l){15-17} 

&&$\mathtt{Inform} \uparrow$ & $\mathtt{Success} \uparrow$ & $\mathtt{BLEU} \uparrow$ & 
$\mathtt{Inform} \uparrow$ & $\mathtt{Success} \uparrow$ & $\mathtt{BLEU} \uparrow$ & 
$\mathtt{Inform} \uparrow$ & $\mathtt{Success} \uparrow$ & $\mathtt{BLEU} \uparrow$ & 
$\mathtt{Inform} \uparrow$ & $\mathtt{Success} \uparrow$ & $\mathtt{BLEU} \uparrow$ \\
\midrule

\modelp{} & 50 & 75.00	&	57.00	&	{\bf13.50}	&	81.31	&	72.22	&	{\bf11.79}	&	62.00	&	38.00	&	{\bf10.02}	&	79.00	&	50.00	&	{\bf12.23} \\
\modelp{} & 10 & 45.00	&	19.00	&	7.67	&	67.68	&	58.08	&	7.13	&	33.50	&	22.50	&	8.70	&	50.50	&	10.00	&	8.61 \\
\modelp{MT} & +5$^*$  & 78.00	&	45.00	&	11.90	&	68.18	&	63.64	&	9.45	&	46.50	&	22.50	&	7.68	&	53.00	&	32.00	&	9.81 \\
\model{} & 5 & {\bf89.00} & {\bf77.00} & 6.03 & {\bf79.80} & {\bf74.24} & 7.36 & {\bf83.50} & {\bf76.50} & 5.89 & {\bf82.00} & {\bf67.00} & 6.69 \\
~ ~ ~w/o $\mathtt{MT}$ & 0 & 87.00 & 76.00 & 5.68 & 79.80 & 71.21 & 6.86 & 82.50 & 74.50 & 5.91 & 78.00 & 67.00 & 6.57 \\
    \bottomrule
\multicolumn{8}{l}{\scriptsize $^*$ $\mathtt{MT}$ is conducted after \modelp{} is being fine-tuned with 10 examples.}
    \end{tabular}
    }
    }
    
    \caption{End-to-end evaluation results on RADDLE.}
    \label{table:overall}
    
\end{table*}



\begin{table}[htbp]
    \centering
    \small
    \scalebox{0.8}{
    \setlength{\tabcolsep}{1.0mm}{
    \begin{tabular}{lccccc}
    \toprule
    
\multirow{2}{*}{Model} & \multirow{2}{*}{\#Example} & \multicolumn{4}{c}{Combined Score $\uparrow$ } \\
\cmidrule(l){3-6}
&& \texttt{Attraction} & \texttt{Train} & \texttt{Hotel} & \texttt{Restaurant}\\
\midrule
\modelp{} & 50 & 79.50 & {\bf88.56} & 60.02 & 76.73 \\
\modelp{} & 10 & 39.67 & 70.01 & 36.70 & 38.86 \\
 \modelp{MT} & +5$^*$ & 73.40 & 75.36 & 42.18 & 52.31 \\
\model{} & 5 & {\bf89.03}	&	84.39	&	{\bf85.88}	&	{\bf81.18} \\
 ~ ~ ~w/o $\mathtt{MT}$ & 0 &87.18	&	82.36	&	84.40	&	79.07 \\
    \bottomrule
\multicolumn{6}{l}{\scriptsize $^*$ $\mathtt{MT}$ is conducted after \modelp{} is fine-tuned with 10 examples.}
    \end{tabular}
    }
    }
    \caption{Combined scores of different methods.}
    \label{table:overall_combined}
\end{table}

\begin{table}[htbp]
    \centering
    \small
    \scalebox{0.8}{
    \setlength{\tabcolsep}{1.0mm}{
    \begin{tabular}{lccccc}
    \toprule
    
\multirow{2}{*}{Model} & \multirow{2}{*}{\#Example} & \multicolumn{4}{c}{Joint Goal Accuracy $\uparrow$ } \\
\cmidrule(l){3-6}
&& \texttt{Attraction} & \texttt{Train} & \texttt{Hotel} & \texttt{Restaurant}\\
\midrule

\modelp{} & 50 & 56.10 & \textbf{60.47} & 29.10 & 67.01 \\
\modelp{} & 10 & 22.85 & 26.25 & 8.24 & 29.56 \\
\modelp{MT}  & +5$^*$ & 42.33 & 29.49 & 14.17 & 38.80\\
\model{} & 5 & \textbf{69.87} & 55.26 & 40.31 & \textbf{66.78} \\
~ ~ ~ w/o $\mathtt{MT}$ & 0 & \textbf{69.87} & 55.36 & \textbf{41.80} & 66.55 \\
    \bottomrule
\multicolumn{6}{l}{\scriptsize $^*$ $\mathtt{MT}$ is conducted after \modelp{} is being fine-tuned with 10 examples.}
    \end{tabular}
    }
    }
    \caption{Dialog state tracking evaluation results.}
    \label{table:dst}
\end{table}

\section{Experiments}

In this section, we evaluate the proposed \model{} in light of two research questions: \textbf{\texttt{Q1}}: How effective is symbolic knowledge used to build neural dialog models for task completion? \textbf{\texttt{Q2}}: How effective is machine teaching for neural learning?

\subsection{Experimental Setup}

\paragraph{Dataset.} We validate the end-to-end dialog system performance of \model{} on RADDLE \cite{DBLP:journals/corr/abs-2012-14666}, which is a subset of MultiWOZ for few-shot end-to-end dialog model evaluation. It contains four domains and each domain has 50/200 dialogs for training and testing except that Attraction has 100 dialogs for testing.

\paragraph{Automatic Evaluation Metrics.} Following \citet{budzianowski2018multiwoz} and \citet{DBLP:journals/corr/abs-2012-14666}, $\mathtt{Inform}$, $\mathtt{Success}$, and $\mathtt{BLEU}$ scores are reported. 
The first two metrics concern the dialog task completion -- whether the system has tracked users' goal accurately ($\mathtt{Inform}$), and then answered all the requested attributes and provided necessary entities ($\mathtt{Success}$). $\mathtt{BLEU}$ assesses how natural the generated responses are compared to that generated by human agents.
A combined score ($\mathtt{Combined}$) is also reported using $\mathtt{Combined} = (\mathtt{Inform} + \mathtt{Success}) \times 0.5 + \mathtt{BLEU}$ as an overall quality measure.

\paragraph{Human Evaluation Metrics.} We follow the same evaluation protocol used by DSTC9 Track 1 challenge \cite{gunasekara2020overview} to conduct human evaluations by employing Amazon Mechanic Turkers to converse with the deployed agent via natural language to fulfill the given user goals. At the end of each dialog session, annotators are asked to assess the dialog quality using the following metrics: 
$(\RN{2})$ $\mathtt{Success}(\%)$ judges whether the agent complete the task and provide matched slot values against the database record. 
$(\RN{3})$ $\mathtt{Understanding}$(1-5) measures the understanding correctness of user utterances. 
$(\RN{4})$ $\mathtt{Appropriateness}$(1-5) indicates the naturalness, appropriateness, and fluency of the response. 
$(\RN{5})$ $\mathtt{Turns}$ is the average number of turns only for successful dialog sessions.

\paragraph{Training.} We leverage the pre-trained SOLOIST checkpoint from \citet{peng2020soloist}, and continually train them on the synthesized dialogs. Each domain is trained separately with a mini-batch of 4 and a learning rate of 5e-5 on 4 Nvidia V100 GPUs until no progress is observed on validation data or up to 10 epochs. 


\subsection{Results and Analysis}


Tables \ref{table:overall} and \ref{table:overall_combined} report the end-to-end evaluation results. Table \ref{table:dst} reports the results of dialog state tracking, which is one of the most important component task for evaluating task-oriented dialog systems.

All the task bots for comparison are based on end-to-end neural dialog models fine-tuned on the pre-trained \modelp{} model \cite{peng2020soloist}, which is publicly available. We compare four types of models.
\begin{enumerate}
    \item \modelp{} are obtained via regular fine-tuning, where a set of task-specific dialogs are randomly sampled from training data (\ie dialog logs) and are fully annotated by humans to fine-tune the pre-trained \modelp.
    \item \modelp{MT} is fine-tuned via machine teaching where a set of failed dialogs are selected from logs using active learning and human teachers label these dialogs by correcting their belief states and responses so that the system can complete the dialog tasks successfully.
    \item \model{} w/o $\mathtt{MT}$ is fine-tuned using only the first step of \model{}, symbolic-knowledge-to-neural-network, where the pre-trained dialog model is fine-tuned using only the simulated dialogs translated from task schema, but not any real dialogs.  
    \item \model{} is fine-tuned the two-step \model{} learning method described in Section 2.2, where the pre-trained \modelp{} is fine-tuned using both the simulated dialogs translated from task schema and real dialogs via machine teaching.
\end{enumerate}

\paragraph{Results of Symbolic-Knowledge-to-Neural-Network.} 
\model{} w/o $\mathtt{MT}$ is fine-tuned only using the simulated dialogs, and thus does not incur any human annotation cost. It achieves substantially better performance than \modelp{MT} fine-tuned using up to 15 human-annotated examples, in terms of $\mathtt{Inform}$ and $\mathtt{Success}$.  
In order for \soloist{} to achieve a comparable performance to that of \model{} w/o $\mathtt{MT}$ 50 human-annotated dialog sessions are needed for fine-tuning. 
This highlights the effectiveness of incorporating symbolic knowledge for neural dialog modeling. 
It is worth mentioning that although \modelp{MT} obtains higher $\mathtt{BLEU}$ scores than \model{} w/o $\mathtt{MT}$, whether $\mathtt{BLEU}$ score is an appropriate metric for task-oriented dialog remains an open question and task success rate ($\mathtt{Success}$) is still widely considered a far more important metric for task bots. 
Therefore, we use $\mathtt{Combined}$ scores listed in Table \ref{table:overall_combined} to measure the overall performance of different models.
We can see that by translating symbolic knowledge (\ie task schema) to neural models, \model{} allows to build high-performance neural dialog models for task completion with little to zero human annotation cost. 

\paragraph{Results of Machine Teaching.}
\citet{peng2020soloist} reported that machine learning is more effective and cost-efficient than regular fine-tuning methods. Our results in Tables \ref{table:overall} and \ref{table:overall_combined} confirm the conclusion of \citet{peng2020soloist}.
We see that machine teaching significantly improves the fine-tuned \soloist{} using only 5 dialog sessions (\soloist{} vs. \modelp{MT}).  
In addition, we see that \model{} significantly outperforms \model{} w/o $\mathtt{MT}$ even if the real dialogs that are fed to the system for fine-tuning via machine teaching (i.e., the neural learning step in \model{}) are of much smaller amounts compared to that of simulated dialogs used in the symbolic-knowledge-to-neural-network step, validating the effectiveness of using neural learning via machine teaching in \model{}. 


\paragraph{Results of Dialog State Tracking (DST).} 
A dialog state is represented in the form of slot-value pairs, and is a summary of the dialog till the current turn. It comprises all information that is needed for the dialog system to not only decide what to action to take next but also form a query to retrieve entities for task-specific databases. The task of DST is to generate a distribution over all possible slot-value pairs at each dialog turn. DST is the one of most important component tasks used to evaluate the performance of a task bot.
We report the DST results of different end-to-end dialog models in Table \ref{table:dst}. Benefiting from enumerating data records to cover all possible slot values, \model{} achieves superior performance with a considerably less amount of annotation efforts.

\begin{figure*}[t!]
\centering
\includegraphics[width=2\columnwidth]{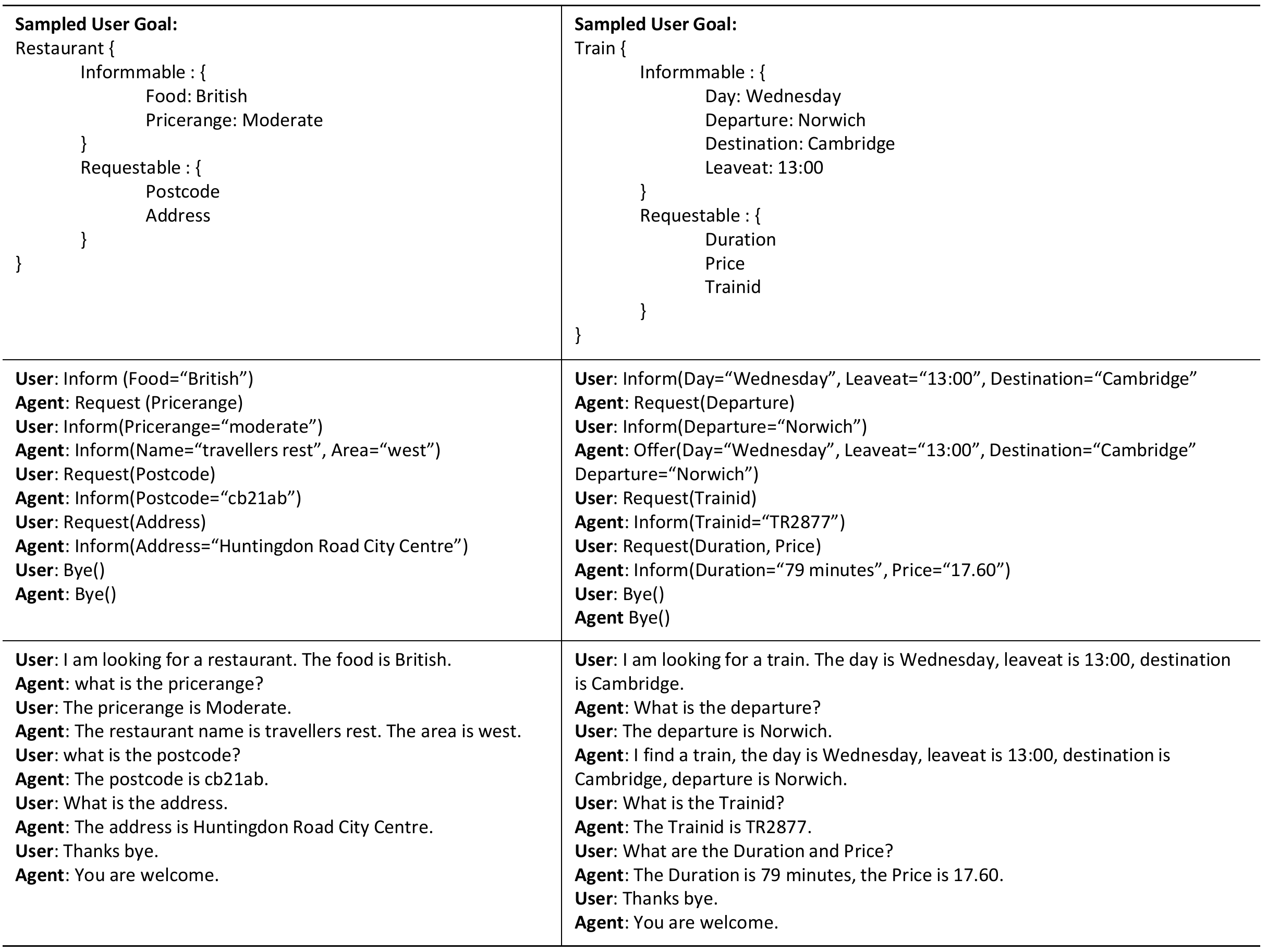}
\caption{Simulated dialog sessions from \texttt{Restaurant} and \texttt{Train} domains in both dialog act (middle row) and natural language level (bottom row). }
\label{fig:sample}
\end{figure*}

Figure \ref{fig:sample} depicts simulated dialog sessions, in the forms of both dialog act and natural language, which are generated from task schema. Although experienced dialog developers can easily distinguish them from real dialogs due to \eg their lack of diversity in language style, the simulated dialogs turn out to be an effective way of encoding task-specific symbolic knowledge (\ie task schema) for task-oriented neural dialog model training.


\begin{table}[htbp]
    \centering
    \small
    \scalebox{0.8}{
    \setlength{\tabcolsep}{1.0mm}{
    \begin{tabular}{lccccc}
    \toprule
    
\multirow{2}{*}{Model} & \multirow{2}{*}{\#Example} & \multicolumn{4}{c}{Restaurant} \\
\cmidrule(l){3-6}
&& \texttt{Success} $\uparrow$ & \texttt{Under.} $\uparrow$ & \texttt{Appr.} $\uparrow$ & \texttt{Turns} $\downarrow$\\
\midrule

\modelp{MT}  & +5$^*$ & 18.00 & 3.91 & 3.82 & 11.94 \\
\model{} & 5 & \textbf{30.00} & \textbf{4.50} & \textbf{4.32} & \textbf{10.17} \\

    \bottomrule
\multicolumn{6}{l}{\scriptsize $^*$ $\mathtt{MT}$ is conducted after \modelp{} is being fine-tuned with 10 examples.}
    \end{tabular}
    }
    }
    \caption{Human evaluation results. Under.: Understanding score, Appr.: Appropriateness score.}
    \label{table:human_eavlue}
\end{table}

\paragraph{Results of Interactive Human Evaluation.} 
Corpus-based automatic evaluation metrics sometimes can not adequately reflect the capability of dialog systems for helping users in real-world scenarios. Hence, we conduct human evaluations on \texttt{Restaurant} domain to evaluate the performance of \modelp{MT} and \model{} by interacting with real users. We gathered 50 dialogs for each system for analysis.

Table \ref{table:human_eavlue} lists the results. Our proposed \model{} performs significantly better than  \modelp{MT}, which is consistent with automatic evaluation. An intriguing observation is that although \modelp{MT} achieve better BLEU score than \model{}, real human users rate \model{} much higher than \modelp{MT} in terms of understanding and appropriate score. 

\section{Related Work}

\paragraph{End-to-End Dialog Systems.} Neural models have shown dominant performance for end-to-end dialog systems \cite{wen2016network,li2017end,lei2018sequicity,haotian2019end}. Although achieving promising results, these methods requires a large number of domain-specific training examples, which is costly to collect. Recently, several studies have explored pre-trained models to improve model generalization and reduce training exampled needed \cite{devlin2019bert, bao2019plato, Wu2020ToDBERTPN, DBLP:conf/emnlp/HendersonCMSWV20, peng2020few, DBLP:conf/acl/CoopeFGVH20,peng2020soloist}. Another line of research to reduce labeling cost of building dialog systems is to automatically create larger datasets from small samples. Such methods involve paraphrasing utterance of existing corpus to augmenting training samples using generative models \cite{hou2018sequence,gao2020paraphrase}. These methods increase language variety but fail to bring new knowledge to a task. In contrast, our work incorporates symbolic knowledge based on task schema and database for task-completion, and is complementary to above methods.


\paragraph{Relation to M2M.} The most related work to ours is the idea of dataset creation via {\it machine-to-machine} conversation simulation dating back to work by~\citet{bordes2016learning}, where pre-defined rules are utilized to expand key information into template-based utterance. It has been revisited recently with the {\it machines talking to machines (M2M)} proposal of ~\citet{shah2018building}. In M2M, crowd workers are employed to further paraphrase the template-generated utterances to natural language, and the paraphrase data is used to train a modular dialog system. Our work is different from M2M in that $(\RN{1})$ unnecessary to hire crowd workers to conduct paraphrasing owning to the employment of pre-trained dialog models; 
$(\RN{2})$ we adopt machine teaching to continue to refine dialog models with a handful of examples.

\paragraph{Symbolic Knowledge and Neural Networks.} Incorporating symbolic knowledge into neural networks has been a long-standing research topic \cite{shavlik1994combining}. It has attracted considerable attention in AI. For example, \citet{towell1994knowledge} introduced rule-to-network translator to map task-specific rules to neural network weights. \citet{xie2019embedding} proposed to use Graph Convolutional Networks to encode symbolic knowledge as a semantic regularization to guide neural networks training. In the natural language processing community, several works have explored leveraging knowledge for learning contextual word representations \cite{peters2019knowledge,wang2020k}, open-domain response generation \cite{DBLP:conf/aaai/GhazvininejadBC18}, commonsense reasoning \cite{xu2020fusing}, etc. To the best of our knowledge, this paper is the first to explicitly exploit symbolic knowledge for building neural dialog models for task completion.

\section{Conclusions}

In this paper, we present the first hybrid learning framework \model{}, which incorporates symbolic knowledge and machine teaching to build task bots. \model{} injects symbolic knowledge encoded in dialog flows and task-specific databases into neural dialog models by training on simulated dialogs. Machine teaching is utilized to continually refine neural models with a handful of real dialogs. We evaluate our method on four non-trivial dialog tasks. The results demonstrate that injecting symbolic knowledge into neural dialog models can significantly improve task-completion performance, and machine teaching is an effective approach to refining dialog models. In summary, we show strong synergies 
between symbolic knowledge and neural models. 

\bibliography{aaai2022}

\begin{thebibliography}{38}
\expandafter\ifx\csname natexlab\endcsname\relax\def\natexlab#1{#1}\fi

\bibitem[{Bao et~al.(2019)Bao, He, Wang, and Wu}]{bao2019plato}
Siqi Bao, Huang He, Fan Wang, and Hua Wu. 2019.
\newblock Plato: Pre-trained dialogue generation model with discrete latent
  variable.
\newblock \emph{arXiv preprint arXiv:1910.07931}.

\bibitem[{Bordes et~al.(2016)Bordes, Boureau, and Weston}]{bordes2016learning}
Antoine Bordes, Y-Lan Boureau, and Jason Weston. 2016.
\newblock Learning end-to-end goal-oriented dialog.
\newblock \emph{arXiv preprint arXiv:1605.07683}.

\bibitem[{Budzianowski et~al.(2018)Budzianowski, Wen, Tseng, Casanueva, Ultes,
  Ramadan, and Ga{\v{s}}i{\'c}}]{budzianowski2018multiwoz}
Pawe{\l} Budzianowski, Tsung-Hsien Wen, Bo-Hsiang Tseng, Inigo Casanueva,
  Stefan Ultes, Osman Ramadan, and Milica Ga{\v{s}}i{\'c}. 2018.
\newblock Multiwoz - a large-scale multi-domain wizard-of-oz dataset for
  task-oriented dialogue modelling.
\newblock \emph{arXiv preprint arXiv:1810.00278}.

\bibitem[{Coope et~al.(2020)Coope, Farghly, Gerz, Vulic, and
  Henderson}]{DBLP:conf/acl/CoopeFGVH20}
Sam Coope, Tyler Farghly, Daniela Gerz, Ivan Vulic, and Matthew Henderson.
  2020.
\newblock \href {https://www.aclweb.org/anthology/2020.acl-main.11/}
  {Span-convert: Few-shot span extraction for dialog with pretrained
  conversational representations}.
\newblock In \emph{Proceedings of the 58th Annual Meeting of the Association
  for Computational Linguistics, {ACL} 2020, Online, July 5-10, 2020}, pages
  107--121. Association for Computational Linguistics.

\bibitem[{Devlin et~al.(2019)Devlin, Chang, Lee, and
  Toutanova}]{devlin2019bert}
Jacob Devlin, Ming-Wei Chang, Kenton Lee, and Kristina Toutanova. 2019.
\newblock {BERT}: Pre-training of deep bidirectional transformers for language
  understanding.
\newblock \emph{NAACL}.

\bibitem[{Eric et~al.(2020)Eric, Goel, Paul, Sethi, Agarwal, Gao, Kumar, Goyal,
  Ku, and Hakkani-Tur}]{eric2020multiwoz}
Mihail Eric, Rahul Goel, Shachi Paul, Abhishek Sethi, Sanchit Agarwal, Shuyang
  Gao, Adarsh Kumar, Anuj Goyal, Peter Ku, and Dilek Hakkani-Tur. 2020.
\newblock Multiwoz 2.1: A consolidated multi-domain dialogue dataset with state
  corrections and state tracking baselines.
\newblock In \emph{Proceedings of The 12th Language Resources and Evaluation
  Conference}, pages 422--428.

\bibitem[{Gao et~al.(2019{\natexlab{a}})Gao, Galley, and Li}]{gao2019neural}
Jianfeng Gao, Michel Galley, and Lihong Li. 2019{\natexlab{a}}.
\newblock Neural approaches to conversational ai.
\newblock \emph{Foundations and Trends{\textregistered} in Information
  Retrieval}, 13(2-3):127--298.

\bibitem[{Gao et~al.(2020{\natexlab{a}})Gao, Peng, Li, Li, Shayandeh, Liden,
  and Shum}]{gao2020robust}
Jianfeng Gao, Baolin Peng, Chunyuan Li, Jinchao Li, Shahin Shayandeh, Lars
  Liden, and Heung{-}Yeung Shum. 2020{\natexlab{a}}.
\newblock \href {http://arxiv.org/abs/2009.03457} {Robust conversational {AI}
  with grounded text generation}.
\newblock \emph{CoRR}, abs/2009.03457.

\bibitem[{Gao et~al.(2019{\natexlab{b}})Gao, Sethi, Agarwal, Chung, and
  Hakkani-Tur}]{gao2019dialog}
Shuyang Gao, Abhishek Sethi, Sanchit Agarwal, Tagyoung Chung, and Dilek
  Hakkani-Tur. 2019{\natexlab{b}}.
\newblock Dialog state tracking: A neural reading comprehension approach.
\newblock In \emph{Proceedings of the 20th Annual SIGdial Meeting on Discourse
  and Dialogue}, pages 264--273.

\bibitem[{Gao et~al.(2020{\natexlab{b}})Gao, Zhang, Ou, and
  Yu}]{gao2020paraphrase}
Silin Gao, Yichi Zhang, Zhijian Ou, and Zhou Yu. 2020{\natexlab{b}}.
\newblock Paraphrase augmented task-oriented dialog generation.
\newblock In \emph{Proceedings of the 58th Annual Meeting of the Association
  for Computational Linguistics}, pages 639--649.

\bibitem[{Ghazvininejad et~al.(2018)Ghazvininejad, Brockett, Chang, Dolan, Gao,
  Yih, and Galley}]{DBLP:conf/aaai/GhazvininejadBC18}
Marjan Ghazvininejad, Chris Brockett, Ming{-}Wei Chang, Bill Dolan, Jianfeng
  Gao, Wen{-}tau Yih, and Michel Galley. 2018.
\newblock \href
  {https://www.aaai.org/ocs/index.php/AAAI/AAAI18/paper/view/16710} {A
  knowledge-grounded neural conversation model}.
\newblock In \emph{Proceedings of the Thirty-Second {AAAI} Conference on
  Artificial Intelligence, (AAAI-18), the 30th innovative Applications of
  Artificial Intelligence (IAAI-18), and the 8th {AAAI} Symposium on
  Educational Advances in Artificial Intelligence (EAAI-18), New Orleans,
  Louisiana, USA, February 2-7, 2018}, pages 5110--5117. {AAAI} Press.

\bibitem[{Goel et~al.(2019)Goel, Paul, and Hakkani-T{\"u}r}]{goel2019hyst}
Rahul Goel, Shachi Paul, and Dilek Hakkani-T{\"u}r. 2019.
\newblock Hyst: A hybrid approach for flexible and accurate dialogue state
  tracking.
\newblock \emph{Proc. Interspeech 2019}, pages 1458--1462.

\bibitem[{Gunasekara et~al.(2020)Gunasekara, Kim, D'Haro, Rastogi, Chen, Eric,
  Hedayatnia, Gopalakrishnan, Liu, Huang et~al.}]{gunasekara2020overview}
Chulaka Gunasekara, Seokhwan Kim, Luis~Fernando D'Haro, Abhinav Rastogi,
  Yun-Nung Chen, Mihail Eric, Behnam Hedayatnia, Karthik Gopalakrishnan, Yang
  Liu, Chao-Wei Huang, et~al. 2020.
\newblock Overview of the ninth dialog system technology challenge: Dstc9.
\newblock \emph{arXiv preprint arXiv:2011.06486}.

\bibitem[{Haotian et~al.(2019)Haotian, Haiyun, Haoran, CAMBRIA, Liuyang, and
  ZHENG}]{haotian2019end}
XU~Haotian, PENG Haiyun, XIE Haoran, Erik CAMBRIA, ZHOU Liuyang, and Weiguo
  ZHENG. 2019.
\newblock End-to-end latent-variable task-oriented dialogue system with exact
  log-likelihood optimization.
\newblock \emph{World Wide Web}, pages 1--14.

\bibitem[{Henderson et~al.(2020)Henderson, Casanueva, Mrksic, Su, Wen, and
  Vulic}]{DBLP:conf/emnlp/HendersonCMSWV20}
Matthew Henderson, I{\~{n}}igo Casanueva, Nikola Mrksic, Pei{-}Hao Su,
  Tsung{-}Hsien Wen, and Ivan Vulic. 2020.
\newblock \href {https://www.aclweb.org/anthology/2020.findings-emnlp.196/}
  {Convert: Efficient and accurate conversational representations from
  transformers}.
\newblock In \emph{Proceedings of the 2020 Conference on Empirical Methods in
  Natural Language Processing: Findings, {EMNLP} 2020, Online Event, 16-20
  November 2020}, pages 2161--2174. Association for Computational Linguistics.

\bibitem[{Hosseini-Asl et~al.(2020)Hosseini-Asl, McCann, Wu, Yavuz, and
  Socher}]{hosseini2020simple}
Ehsan Hosseini-Asl, Bryan McCann, Chien-Sheng Wu, Semih Yavuz, and Richard
  Socher. 2020.
\newblock A simple language model for task-oriented dialogue.
\newblock \emph{arXiv preprint arXiv:2005.00796}.

\bibitem[{Hou et~al.(2018)Hou, Liu, Che, and Liu}]{hou2018sequence}
Yutai Hou, Yijia Liu, Wanxiang Che, and Ting Liu. 2018.
\newblock Sequence-to-sequence data augmentation for dialogue language
  understanding.
\newblock In \emph{Proceedings of the 27th International Conference on
  Computational Linguistics}, pages 1234--1245.

\bibitem[{Kambhampati(2021)}]{kambhampati2021polanyi}
Subbarao Kambhampati. 2021.
\newblock Polanyi's revenge and ai's new romance with tacit knowledge.
\newblock \emph{Communications of the ACM}.

\bibitem[{Lei et~al.(2018)Lei, Jin, Kan, Ren, He, and Yin}]{lei2018sequicity}
Wenqiang Lei, Xisen Jin, Min-Yen Kan, Zhaochun Ren, Xiangnan He, and Dawei Yin.
  2018.
\newblock Sequicity: Simplifying task-oriented dialogue systems with single
  sequence-to-sequence architectures.
\newblock In \emph{Proceedings of the 56th Annual Meeting of the Association
  for Computational Linguistics (Volume 1: Long Papers)}.

\bibitem[{Li et~al.(2017)Li, Chen, Li, Gao, and Celikyilmaz}]{li2017end}
Xiujun Li, Yun-Nung Chen, Lihong Li, Jianfeng Gao, and Asli Celikyilmaz. 2017.
\newblock End-to-end task-completion neural dialogue systems.
\newblock \emph{arXiv preprint arXiv:1703.01008}.

\bibitem[{Peng et~al.(2021)Peng, Li, Li, Shayandeh, Liden, and
  Gao}]{peng2020soloist}
Baolin Peng, Chunyuan Li, Jinchao Li, Shahin Shayandeh, Lars Liden, and
  Jianfeng Gao. 2021.
\newblock Soloist: Building task bots at scale with transfer learning and
  machine teaching.
\newblock \emph{Transactions of the Association for Computational Linguistics},
  9:807--824.

\bibitem[{Peng et~al.(2020{\natexlab{a}})Peng, Li, Zhang, Zhu, Li, and
  Gao}]{DBLP:journals/corr/abs-2012-14666}
Baolin Peng, Chunyuan Li, Zhu Zhang, Chenguang Zhu, Jinchao Li, and Jianfeng
  Gao. 2020{\natexlab{a}}.
\newblock \href {http://arxiv.org/abs/2012.14666} {{RADDLE:} an evaluation
  benchmark and analysis platform for robust task-oriented dialog systems}.
\newblock \emph{CoRR}, abs/2012.14666.

\bibitem[{Peng et~al.(2017)Peng, Li, Li, Gao, Celikyilmaz, Lee, and
  Wong}]{peng2017composite}
Baolin Peng, Xiujun Li, Lihong Li, Jianfeng Gao, Asli Celikyilmaz, Sungjin Lee,
  and Kam-Fai Wong. 2017.
\newblock \href {https://doi.org/10.18653/v1/D17-1237} {Composite
  task-completion dialogue policy learning via hierarchical deep reinforcement
  learning}.
\newblock In \emph{Proceedings of the 2017 Conference on Empirical Methods in
  Natural Language Processing}, pages 2231--2240, Copenhagen, Denmark.
  Association for Computational Linguistics.

\bibitem[{Peng et~al.(2020{\natexlab{b}})Peng, Zhu, Li, Li, Li, Zeng, and
  Gao}]{peng2020few}
Baolin Peng, Chenguang Zhu, Chunyuan Li, Xiujun Li, Jinchao Li, Michael Zeng,
  and Jianfeng Gao. 2020{\natexlab{b}}.
\newblock \href {https://doi.org/10.18653/v1/2020.findings-emnlp.17} {Few-shot
  natural language generation for task-oriented dialog}.
\newblock In \emph{Findings of the Association for Computational Linguistics:
  EMNLP 2020}, pages 172--182, Online. Association for Computational
  Linguistics.

\bibitem[{Peters et~al.(2019)Peters, Neumann, Logan, Schwartz, Joshi, Singh,
  and Smith}]{peters2019knowledge}
Matthew~E Peters, Mark Neumann, Robert Logan, Roy Schwartz, Vidur Joshi, Sameer
  Singh, and Noah~A Smith. 2019.
\newblock Knowledge enhanced contextual word representations.
\newblock In \emph{Proceedings of the 2019 Conference on Empirical Methods in
  Natural Language Processing and the 9th International Joint Conference on
  Natural Language Processing (EMNLP-IJCNLP)}, pages 43--54.

\bibitem[{Schatzmann et~al.(2007)Schatzmann, Thomson, Weilhammer, Ye, and
  Young}]{schatzmann2007agenda}
Jost Schatzmann, Blaise Thomson, Karl Weilhammer, Hui Ye, and Steve Young.
  2007.
\newblock Agenda-based user simulation for bootstrapping a pomdp dialogue
  system.
\newblock In \emph{Human Language Technologies 2007: The Conference of the
  North American Chapter of the Association for Computational Linguistics;
  Companion Volume, Short Papers}, pages 149--152.

\bibitem[{Shah et~al.(2018)Shah, Hakkani-T{\"u}r, T{\"u}r, Rastogi, Bapna,
  Nayak, and Heck}]{shah2018building}
Pararth Shah, Dilek Hakkani-T{\"u}r, Gokhan T{\"u}r, Abhinav Rastogi, Ankur
  Bapna, Neha Nayak, and Larry Heck. 2018.
\newblock Building a conversational agent overnight with dialogue self-play.
\newblock \emph{arXiv preprint arXiv:1801.04871}.

\bibitem[{Shavlik(1994)}]{shavlik1994combining}
Jude~W Shavlik. 1994.
\newblock Combining symbolic and neural learning.
\newblock \emph{Machine Learning}, 14(3):321--331.

\bibitem[{Shukla et~al.(2020)Shukla, Liden, Shayandeh, Kamal, Li, Mazzola,
  Park, Peng, and Gao}]{shukla2020conversation}
Swadheen Shukla, Lars Liden, Shahin Shayandeh, Eslam Kamal, Jinchao Li, Matt
  Mazzola, Thomas Park, Baolin Peng, and Jianfeng Gao. 2020.
\newblock Conversation learner--a machine teaching tool for building dialog
  managers for task-oriented dialog systems.
\newblock \emph{arXiv preprint arXiv:2004.04305}.

\bibitem[{Towell and Shavlik(1994)}]{towell1994knowledge}
Geoffrey~G Towell and Jude~W Shavlik. 1994.
\newblock Knowledge-based artificial neural networks.
\newblock \emph{Artificial intelligence}, 70(1-2):119--165.

\bibitem[{Wang et~al.(2020)Wang, Tang, Duan, Wei, Huang, Cao, Jiang, Zhou
  et~al.}]{wang2020k}
Ruize Wang, Duyu Tang, Nan Duan, Zhongyu Wei, Xuanjing Huang, Cuihong Cao,
  Daxin Jiang, Ming Zhou, et~al. 2020.
\newblock K-adapter: Infusing knowledge into pre-trained models with adapters.
\newblock \emph{arXiv preprint arXiv:2002.01808}.

\bibitem[{Wen et~al.(2016)Wen, Vandyke, Mrksic, Gasic, Rojas-Barahona, Su,
  Ultes, and Young}]{wen2016network}
Tsung-Hsien Wen, David Vandyke, Nikola Mrksic, Milica Gasic, Lina~M
  Rojas-Barahona, Pei-Hao Su, Stefan Ultes, and Steve Young. 2016.
\newblock A network-based end-to-end trainable task-oriented dialogue system.
\newblock \emph{arXiv preprint arXiv:1604.04562}.

\bibitem[{Williams et~al.(2017)Williams, Asadi, and Zweig}]{williams2017hybrid}
Jason~D Williams, Kavosh Asadi, and Geoffrey Zweig. 2017.
\newblock Hybrid code networks: practical and efficient end-to-end dialog
  control with supervised and reinforcement learning.
\newblock \emph{arXiv preprint arXiv:1702.03274}.

\bibitem[{Wu et~al.(2020)Wu, Hoi, Socher, and Xiong}]{Wu2020ToDBERTPN}
Chien-Sheng Wu, Steven Hoi, Richard Socher, and Caiming Xiong. 2020.
\newblock {ToD}-{BERT}: Pre-trained natural language understanding for
  task-oriented dialogues.

\bibitem[{Xie et~al.(2019)Xie, Xu, Meel, Kankanhalli, and
  Soh}]{xie2019embedding}
Yaqi Xie, Ziwei Xu, Kuldeep~S Meel, Mohan~S Kankanhalli, and Harold Soh. 2019.
\newblock Embedding symbolic knowledge into deep networks.
\newblock In \emph{NeurIPS}.

\bibitem[{Xu et~al.(2020)Xu, Zhu, Xu, Liu, Zeng, and Huang}]{xu2020fusing}
Yichong Xu, Chenguang Zhu, Ruochen Xu, Yang Liu, Michael Zeng, and Xuedong
  Huang. 2020.
\newblock \href {http://arxiv.org/abs/2012.04808} {Fusing context into
  knowledge graph for commonsense reasoning}.

\bibitem[{Young et~al.(2013)Young, Gasic, Thomson, and
  Williams}]{DBLP:journals/pieee/YoungGTW13}
Steve~J. Young, Milica Gasic, Blaise Thomson, and Jason~D. Williams. 2013.
\newblock \href {https://doi.org/10.1109/JPROC.2012.2225812} {Pomdp-based
  statistical spoken dialog systems: {A} review}.
\newblock \emph{Proc. {IEEE}}, 101(5):1160--1179.

\bibitem[{Zhou et~al.()Zhou, Gao, Li, and Shum}]{zhou2018design}
Li~Zhou, Jianfeng Gao, Di~Li, and Heung-Yeung Shum.
\newblock The design and implementation of xiaoice, an empathetic social
  chatbot.
\newblock \emph{Computational Linguistics}, (Just Accepted):1--62.

\end{thebibliography}
\bibliographystyle{aaai}
\end{document}